\documentclass[a4paper,fleqn]{cas-dc}
\usepackage[numbers]{natbib}
\usepackage{times}
\usepackage{epsfig}
\usepackage{graphicx}
\usepackage{amsmath}
\usepackage{amssymb}
\usepackage{makecell}
\usepackage{bigstrut,multirow,rotating}
\usepackage{algorithm}
\usepackage{algorithmic}
\usepackage{hyperref}

\def\tsc#1{\csdef{#1}{\textsc{\lowercase{#1}}\xspace}}
\tsc{WGM}
\tsc{QE}

\begin{document}
\let\WriteBookmarks\relax
\def\floatpagepagefraction{1}
\def\textpagefraction{.001}
\let\printorcid\relax 

\shorttitle{}    


\title[mode = title]{Adept: Annotation-Denoising Auxiliary Tasks with Discrete Cosine Transform Map and Keypoint for Human-Centric Pretraining}  


\author[1]{Weizhen He}
\ead{hewz@zju.edu.cn}

\author[2,4]{Yunfeng Yan}
\cormark[1]
\ead{yvonnech@zju.edu.cn}

\author[5]{Shixiang Tang}
\ead{tangshixiang@pjlab.org.cn}

\author[1]{Yiheng Deng}
\ead{dengyiheng@zju.edu.cn}

\author[3,4]{Yangyang Zhong}
\ead{12334127@zju.edu.cn}

\author[3]{Pengxin Luo}
\ead{22334038@zju.edu.cn}

\author[1,4]{Donglian Qi}
\ead{qidl@zju.edu.cn}


\address[1]{College of Electrical Engineering, Zhejiang University, Hangzhou, Zhejiang 310027, China}
    
\address[2]{{School of Mechanical Engineering, Zhejiang University, Hangzhou, Zhejiang 310000, China}}
    
\address[3]{Ocean College, Zhejiang University Zhoushan, Zhejiang 316021, China}
    
\address[4]{Hainan Institute, Zhejiang University, Sanya, Hainan 572025, China}

\address[5]{Shanghai AI Laboratory, Shanghai, 200232, China}
    
\cortext[1]{Corresponding author}  

\tnotetext[1]{This work is jointly supported by the National Natural Science Foundation of China (No.62127803), Key R\&D Project of Zhejiang Province (No.2022C01056).}

\begin{abstract}
Human-centric perception is the core of diverse computer vision tasks and has been a long-standing research focus. However, previous research studied these human-centric tasks individually, whose performance is largely limited to the size of the public task-specific datasets. Recent human-centric methods leverage the additional modalities, e.g., depth, to learn fine-grained semantic information, which limits the benefit of pretraining models due to their sensitivity to camera views and the scarcity of RGB-D data on the Internet. This paper improves the data scalability of human-centric pretraining methods by discarding depth information and exploring semantic information of RGB images in the frequency space by Discrete Cosine Transform (DCT). We further propose new annotation denoising auxiliary tasks with keypoints and DCT maps to enforce the RGB image extractor to learn fine-grained semantic information of human bodies. Our extensive experiments show that when pretrained on large-scale datasets (COCO and AIC datasets) without depth annotation, our model achieves better performance than state-of-the-art methods by \textbf{+0.5 mAP}($\uparrow$) on COCO, \textbf{+1.4 PCKh}($\uparrow$) on MPII and \textbf{-0.51 EPE}($\downarrow$) on Human3.6M for pose estimation, by \textbf{+4.50 mIoU}($\uparrow$) on Human3.6M for human parsing, by \textbf{-3.14 MAE}($\downarrow$) on SHA and \textbf{-0.07 MAE}($\downarrow$) on SHB for crowd counting, by \textbf{+1.1 F1 score}($\uparrow$) on SHA and \textbf{+0.8 F1 score}($\uparrow$) on SHA for crowd localization, and by \textbf{+0.1 mAP}($\uparrow$) on Market1501 and \textbf{+0.8 mAP}($\uparrow$) on MSMT for person ReID. We also validate the effectiveness of our method on MPII+NTURGBD datasets. 
\end{abstract}



\begin{keywords}
Computer Vision \sep Human-centric perception \sep  Model Pre-training \sep Discrete Cosine Transform
\end{keywords}

\maketitle

\section{Introduction}
Human-centric perception plays an important role in both computer vision and computer graphics, such as social surveillance and the metaverse. Academically, the concept of human-centric perception relates to various research tasks, ranging from global recognition tasks, e.g., person re-identification (ReID)~\cite{he2021transreid,zhu2020identity,dai2022cluster,wu2023camera,thanh2024enhancing}, to local prediction tasks (e.g., pose estimation~\cite{cheng2020higherhrnet,liu2020improving,yuan2021hrformer,xu2022vitpose, xu2021sunnet, yuan2024multi}, crowd location~\cite{song2021rethinking,liang2022end, nguyen2020saliency}, human parsing~\cite{gong2018instance,gong2017look, hao2021context,su2023boundary}). Although these human-centric perception tasks have their own relevant semantic information to focus on, those semantics all rely on the basic structure of human bodies, which inspires us to explore a data-efficient human-centric pretraining method to benefit the diverse human-centric tasks with a large number of human-centric images in public datasets.

\begin{figure}
    \centering
    \includegraphics[width=\linewidth]{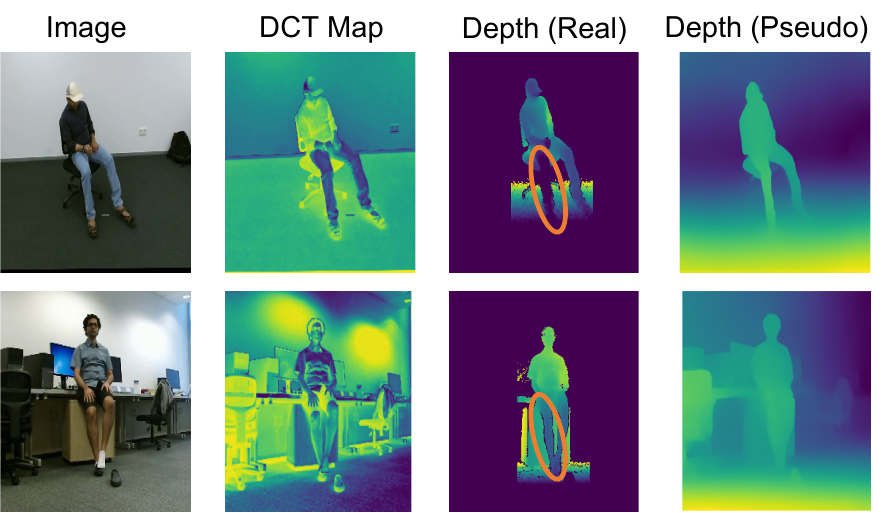}
    \caption{Comparison of fine-grained semantic information in DCT maps, real depth maps captured by RGB-D cameras, and pseudo-depth maps labeled by~\cite{ranftl2021vision}. DCT map can be generated from images with any external sensors or pretrainied models and contains more fine-grained semantic information. The real depth maps can also show fine-grained information, to some extent. The pseudo-depth maps can be easily generated by external models but it contains little fine-grained semantic information. }
    \label{fig:teaser}
\end{figure}

Different from pretraining~\cite{grill2020bootstrap,chen2020simple,he2022masked,bachmann2022multimae,wang2021pfwnet} on natural images where visual appearance varies largely, the challenge of human-centric pretraining lies in perceiving the fine-grained semantic information of human bodies. Existing work~\cite{hong2022versatile} relies on the depth information within RGB-D data to provide semantic information, which limits the data scalability of pretraining methods for the following two reasons. First, compared with RGB-D images that are mostly captured in autonomous driving and indoor scenes, RGB images can be extensively and easily acquired from the Internet with much more diversities outdoor scenes. Second, the benefit of pseudo-labeling depth maps may not necessarily be beneficial. As shown in Fig.~\ref{fig:teaser}, the depth map captured by real cameras can provide valuable fine-grained semantic information (circled by orange) while pseudo-labeled depth maps can not. The high sensitivity for the quality of depth maps plays a big bottleneck in learning fine-grained semantic information about human bodies. 

\begin{figure*}
    \centering
    \includegraphics[width=\linewidth]{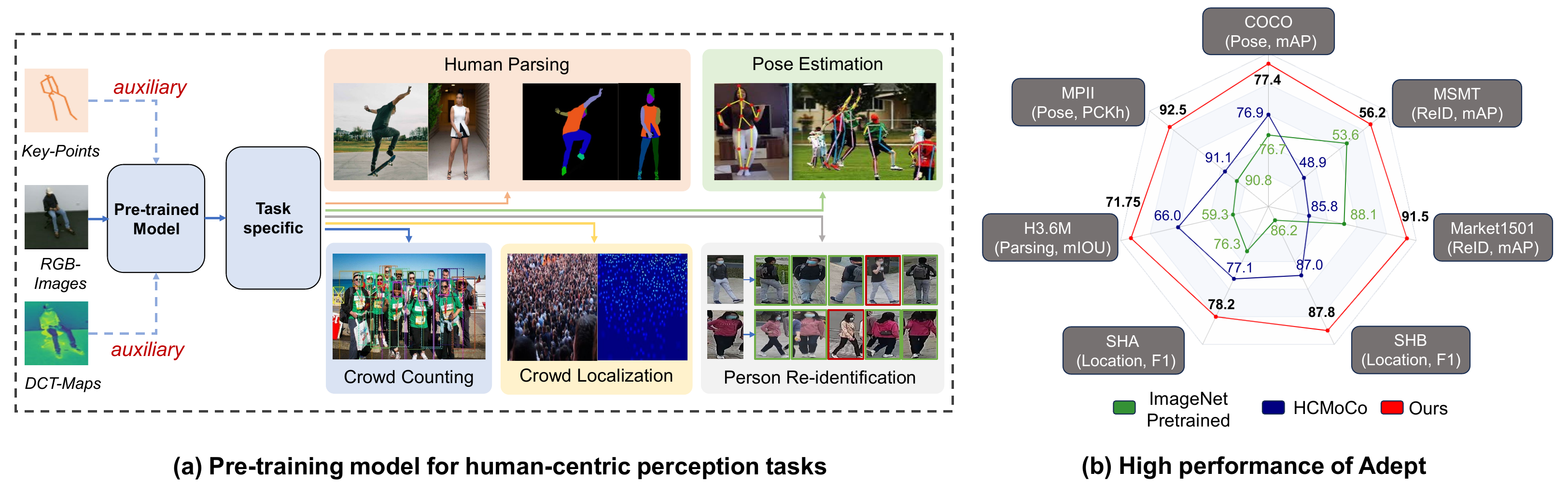}
    \caption{(a) Overview of our proposed Adept framework. Adept first utilizes the Discrete Cosine Transform maps and human key points priors for human-centric pre-training. In the data input, we use dashed lines to represent annotation-denoising auxiliary tasks. Adept then applies the pre-trained model as the backbone for various human-centric perception tasks, \emph{e.g.,} Human Parsing, Pose Estimation, Crowd Counting, Crowd Localization and Person Re-identification. (b) Comparison of ImageNet pretraining and different pretraining methods. Our proposed method improves the performance of various human-centric tasks by a considerable margin.}
    \label{fig:teaser_performance}
\end{figure*}

Therefore, a better fine-grained representation without external pretrained models or additional modalities captured by sensors is required to assist pretraining. In this paper, as shown in Fig.~\ref{fig:teaser_performance} (a), we propose to explore fine-grained semantic information from the frequency domain. The discrete cosine transform (DCT)~\cite{liu2002efficient} is widely used for signal processing and data compression due to its strong ``energy compaction'' property~\cite{rao2014discrete}. 
Regarding RGB images, most semantic information is concentrated in a few low-frequency components. By transforming RGB images into the frequency domain and emphasizing their low-frequency components, we obtain a rough segmentation representation (dubbed as DCT maps in Fig.~\ref{fig:teaser}), which is self-generated and contains more semantic information than depth maps. 

To better learn fine-grained semantic information of human parts, we further propose Adept, \emph{i.e.,} \underline{\textbf{a}}nnotation-\underline{\textbf{de}}noising auxiliary tasks with previously key\underline{\textbf{p}}oints and DC\underline{\textbf{T}} maps. Specifically, we integrate considerable random noise into latent features of annotations and expect the model to reconstruct the clean annotations from these noisy latent features with the assistance of features extracted from RGB images. With this novel design, our annotation-denoising auxiliary tasks can enforce the image backbone to learn fine-grained information because it should help to recover the exact local annotation in both DCT map and keypoints. Our annotation-denoising auxiliary tasks are also more effective than most competing methods, \emph{i.e.,} HCMoCo~\cite{hong2022versatile}, because our supervision signals can be back-propagated to all RGB features while HCMoCo only sparsely contrasts between keypoints and RGB features at the positions of keypoints.

In summary, our contributions are three-fold. First, for the first time, we propose to use the DCT  maps to enable data scalability for human-centric pretraining.  Second, we propose the annotation-denoising auxiliary task which leverages the information from RGB images to reconstruct the annotations from noisy latent features to enforce the model to learn fine-grained semantic features of human bodies. Third, we conduct extensive experiments on human perception tasks and demonstrate that our proposed method can significantly improve performance. Specifically, as shown in Fig.~\ref{fig:teaser_performance} (b), our proposed Adept can consistently improve the performance of 10 datasets on 5 downstream tasks, \emph{e.g.,} person re-identification, crowd counting, crowd localization, pose estimation, and human parsing. For example, for pretraining on AIC and COCO datasets without depth data, our Adept improves the recent state-of-the-art methods by \textbf{+0.4} AP($\uparrow$) on COCO, \textbf{+1.3} AP on MPII($\uparrow$) and \textbf{-0.37} EPE($\downarrow$) on Human3.6M for 2d pose estimation. For human parsing, our method improves the state-of-the-art methods by \textbf{+3.89} mIOU($\uparrow$) on Human3.6M.

\section{Related Work}
\subsection{Learning in the Frequency Domain}
The representations in the frequency domain comprise rich patterns for various image understanding tasks, such as image classification~\cite{xu2020learning} and  segmentation~\cite{xu2020learning}. Early study~\cite{yang2020fda} uncovered that in the frequency domain, the phase component largely has high-level semantics of the original signals, while the amplitude component mainly retains low-level statistics. As such, underlying image semantic patterns can be conveniently observed in the frequency representation. Discrete cosine transform is a widely used algorithm, that encodes the spatial domain RGB images into components in the frequency domain. Based on DCT, DCTransformer~\cite{nash2021generating} and DCT-ResNet~\cite{xu2020learning} are proposed to generate discriminative representation from DCT representations by the different prevalent deep models. Inspired by the working mechanism of the human visual system where different frequency components are treated unequally and the compact nature of DCT representations, Method~\cite{zhong2022detecting} and DCT-Mask~\cite{shen2021dct} leverage DCT to extract other clues beyond RGB image in scene images or to represent semantic masks for object detection and instance segmentation, respectively. Our method is essentially different from previous methods because we leverage DCT in unsupervised learning while the above works are applied to supervised learning. By contrasting images and DCT maps, we target the image backbone learning fine-grained semantic information from DCT maps.

\subsection{Human-centric Pretraining}
Human-centric perception has been studied for decades, including person ReID, pose estimation, human parsing, crowd counting, and crowd localization. Dependent on the targeted perception target, previous research developed methods for various tasks separately, but their performance is highly limited to the scarcity of task-specific public datasets. Recent research started to focus on developing human-centric pretraining methods. HCMoCo leveraged both RGB images and their corresponding depth information, but pretrained human-centric representations for only pose estimation and human parsing. LiftedCL\cite{chenliftedcl} lifted the 2d human annotations to 3d space, but it mainly improves 3d human tasks but shows unsatisfactory gains on 2d tasks because this method does not explicitly learn fine-grained semantic human-part information from images. Our method enforces the image backbone to learn fine-grained semantic information from DCT maps. Different from HCMoCo, we do not need depth maps, enabling data scalability. Different from LiftedCL, we design specific annotation-denoising tasks to learn fine-grained information.

\subsection{Auxiliary Tasks in Visual Perceptions}
The auxiliary tasks are referred to as atomic tasks that are of minor interest or even irrelevant to the application. Despite being seemingly unrelated, they are expected to assist in finding a rich and robust representation of the input image in the common part, from which the ultimately desired main tasks profit. Currently, auxiliary tasks are mainly applied in scene understanding~\cite{liebel2018auxiliary,he2025instruct,he2024instruct}, vision-and-language tasks~\cite{kuo2023structure,ye2021auxiliary,hosseinzadeh2021image,he2023unsupervised}, and efficient learning~\cite{fisch2021few,shi2020auxiliary,kung2021efficient,su2020adapting}, but were not applied to human-centric pretraining methods. Our method is based on contrastive learning but eliminates its disadvantages of focusing only on global appearance by adding two denoising auxiliary tasks to learn fine-grained information embedded in DCT maps and keypoints. 
\vspace{-0.5em}
\subsection{Specialist Methods on Human-centric Tasks.} 
Human-centric tasks play a crucial role in real-world applications such as social surveillance, sports analysis, and \emph{etc}. With the advancement of research, numerous frameworks have been developed to address these tasks, \emph{e.g.,} Unihcp~\cite{ci2023unihcp} proposes a unified framework to handle diverse human-centric tasks with task-specific finetuning, Humanbench~\cite{tang2023humanbench} introduces a shared decoder head with a task-guided interpreter to handle five 2D vision tasks. 
To improve model performance on insufficient training data, research~\cite{cao2022towards} adopts a label distribution learning method to overcome the limitations posed by data shortages in the facial pose estimation task. For universal vision models, Clusterfomer~\cite{liang2024clusterfomer} facilitates efficient and transparent feature distribution in a cluster manner, greatly enhancing performance in perception tasks. In terms of optimizing model architecture, ~\cite{lu2023transflow, liu2021densernet,qin2023reformulating} introduce the specific designs for downstream tasks. Specifically, TransFlow~\cite{lu2023transflow} uses spatial self-attention and cross-attention mechanisms to capture global dependencies and recover lost information, DenserNet~\cite{liu2021densernet} integrates multi-scale feature maps to improve image representations for visual localization, HiGraph+~\cite{qin2023reformulating} advances feature clustering by utilizing graph kernels to capture both structural and temporal consistency. Despite the success of previous specialist methods, they focused on extracting multi-granularity features and designing sophisticated network mechanisms. In contrast, our Adept focuses on pretraining tasks and emphasizes the use of appropriate auxiliary tasks along with denoising techniques. By leveraging DCT maps and keypoints without introducing additional data labeling operations, our approach facilitates fine-grained feature learning and further enhances performance across downstream tasks on existing framework methods.


\section{Methodology}
We now introduce our proposed \textbf{Adept}, \emph{i.e.,} \underline{\textbf{A}}nnotation-\underline{\textbf{de}}noising auxiliary task with key\underline{\textbf{p}}oints and DC\underline{\textbf{T}} maps for human-centric pretraining. Our Adept is based on the contrastive learning framework while its innovations lie in two aspects to facilitate both data scalability and effective fine-grained feature learning. First, different from the previous human-centric pretraining method using RGB-D data, our Adept discards the depth modality and learns human-centric semantic information from the self-generated DCT maps, which practically enables large data scalability. Second, the proposed annotation-denoising task enforces the image backbone to extract fine-grained features in order to recover the fine-grained semantic information in both DCT maps and human keypoints. 

\begin{figure*}
    \centering
    \includegraphics[width=\linewidth]{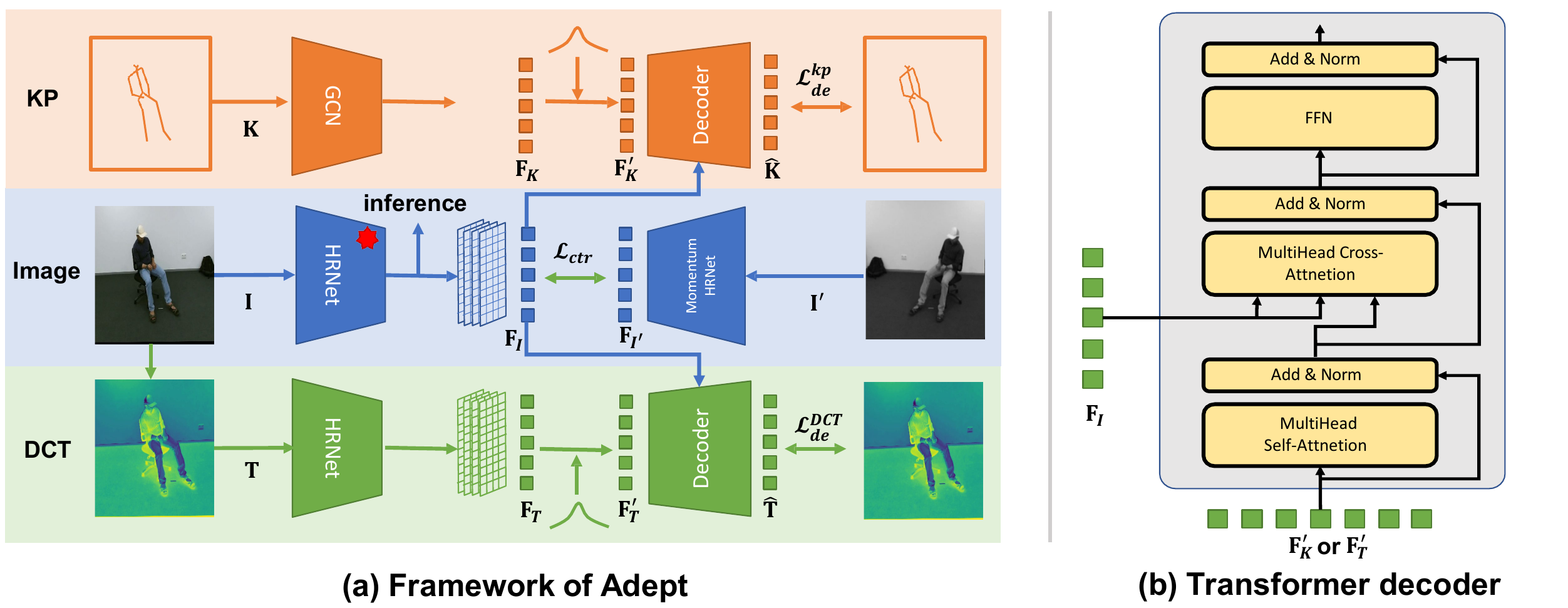}
    \caption{(a) Schematic illustration of our proposed Adept method: Learning local and global features. Given a seed image and keypoint, we first generate the DCT map by Discrete Cosine Transform and two image views. Modality-specific encoders extract latent features of  each modality. Random noises are added to  the latent features of DCT and keypoint, and then task-specific decoders are employed to reconstruct DCT and keypoint by cross-attention with the help of image features, so as to force image encoder to learn fine-grained features.  Additionally, contrastive learning is applied between two image views to extract global features. (b) The schematic illustration of transformer decoders for both keypoints and DCT maps. Specifically, we feed the noisy DCT latent representations and keypoint representations in the multi-head self-attention layer and the latent features of images for cross-attention layer. }
    \label{fig:framework}
\end{figure*}

We illustrate the framework of Adept in Fig.~\ref{fig:framework} (a). The detailed pipeline is described as follows.

\emph{Step1: Generate two different views of an image and its DCT maps (Sec.~\ref{sec:dct})}. Given an image, we get two augmented views $\mathbf{I}$ and $\mathbf{I}'$ and generate the DCT map $\mathbf{T}$ by cosine discrete transformation from $\mathbf{I}$.

\emph{Step2: Extract latent features of RGB images, DCT maps and keypoints by RGB encoder, DCT encoder and keypoint encoder, respectively.} Given two views $\mathbf{I}$, $\mathbf{I}'$, the DCT map $\mathbf{T}$ and keypoints $\mathbf{K}$, we extract their latent representations $\mathbf{F}_I$, $\mathbf{F}_{I'}$, $\mathbf{F}_{T}$ and $\mathbf{F}_K$ by $\mathbf{F}_I=\mathcal{F}_I(\mathbf{I})$, $\mathbf{F}_{I'}=\mathcal{F}_{I}^m(\mathbf{I}')$, $\mathbf{F}_T=\mathcal{F}_T(\mathbf{T})$ and $\mathbf{F}_K=\mathcal{F}_K(\mathbf{K})$, respectively, where $\mathcal{F}_I$, $\mathcal{F}_T$ and $\mathcal{F}_K$ are encoders of RGB images, DCT maps, keypoints and $\mathcal{F}_I^m$ is the momentum encoder for another view of RGB images.

\emph{Step3: Add noise to the latent features of DCT maps and keypoints.}
Given latent features $\mathbf{F}_{T}$ and $\mathbf{F}_K$, we add considerable random noise to them, getting $\mathbf{F}'_{T}$ and $\mathbf{F}'_K$.

\emph{Step4: Recover the DCT maps and keypoints from noisy features with the help of RGB features.} We formulate the noisy features of DCT maps and keypoints as the input of the self-attention layer in the transformer decoder and the latent features of images as the key of cross-attention layers in the transformer decoder to recover the original DCT map and keypoint, respectively, \emph{i.e.,} $\hat{\mathbf{T}}=\mathcal{D}_T(\mathbf{F}'_T, \mathbf{F}_I)$ and $\hat{\mathbf{K}}=\mathcal{D}_K(\mathbf{F}'_K, \mathbf{F}_I)$, where $\mathcal{D}_T$ and $\mathcal{D}_K$ are transformer decoder for DCT map and keypoint.


\emph{Step5: Optimize the image backbone by backward propagation by a summation of the recover loss and contrastive loss ((Eq.~\ref{eq:total_loss})).}

\subsection{Discrete Cosine Transform Map} ~\label{sec:dct}
\begin{figure}
    \centering
    \includegraphics[width=\linewidth]{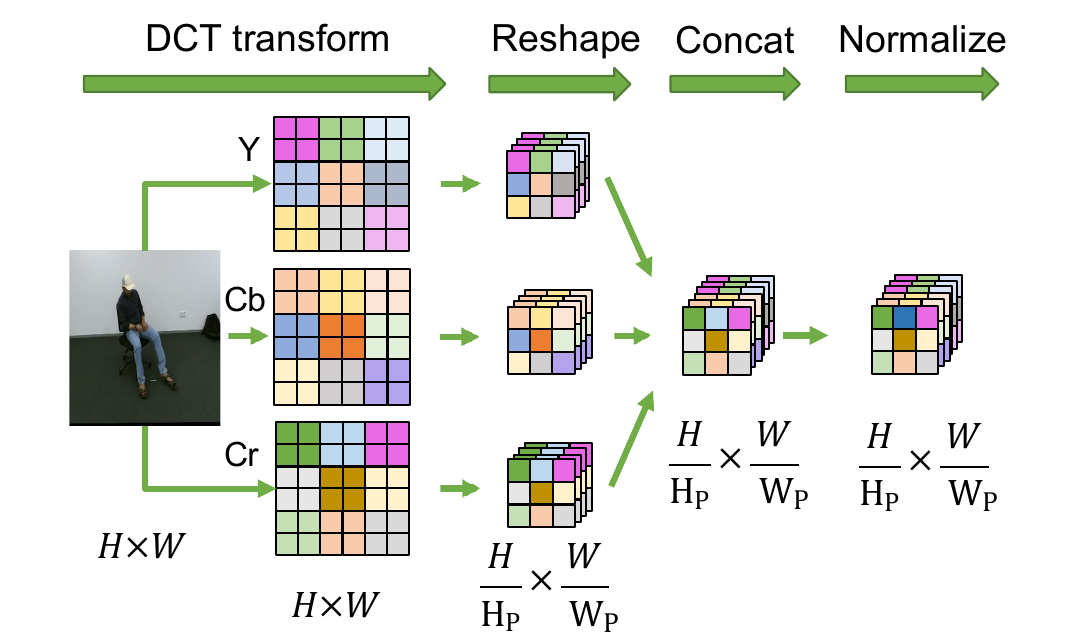}
    \caption{The generation pipeline of Discrete Cosine Transform maps. RGB image is first transformed to the YCbCr color space and then divided to patches for each channel. Discrete Cosine Transform operates on each part to  get DCT coefficients, which are flattened to 1D vector. DCT maps are generated by concatenating DCT coefficients of all channels. }
    \label{fig:dct_generation}
\end{figure}

Compressed representations in the frequency domain contain rich semantic patterns for image understanding. Discrete Cosine Transform can covert images to their frequency space with the strong ``energy compaction'' property. Specifically, we illustrate the generation of a discrete cosine transform map in Fig.~\ref{fig:dct_generation}. Given a view of an image $\mathbf{I}\in \mathbb{R}^{H\times W}$, we first transform it into YCrCb format and divide the whole image into patches $\mathcal{P}=(\mathbf{P}_1, \mathbf{P}_2, ..., \mathbf{P}_N)$ with the size of $H_P \times W_P$, where $N=\frac{H}{H_P} \times \frac{W}{W_P}$ is the number of patches. For every channel $c \in \{Y, Cr, Cb\}$,  we transform the $c$-th channel of $i$-th patch $\mathbf{P}_i$ into frequency space by
\begin{small} 
\begin{equation} \label{eq:dct}
    \footnotesize
    \mathbf{T}^i_{c, u,v} = \frac{1}{4}\alpha_u \alpha_v\sum_{x=0}^{W_P}\sum_{y=0}^{H_P}t_{c, x,y} \cos \left[\frac{(2x+1)u\pi}{2H_p}\right] 
    \cos\left[\frac{(2y+1)v\pi}{2W_p}\right],
\end{equation}
\end{small}
where $\alpha_{u}$ and $\alpha_{v}$ are the normalizing factors, $t_{c, x,y}$ is the pixel value at $(x, y)$ of $c$-th channel, $\mathbf{T}^i_{c,u,v}$ is the DCT coefficient at $(u, v)$ of $c$-th channel and $0 \leq u < H_P, 0 \leq v < W_P$. Afterward, we flatten the 2D DCT features $\mathbf{T}^i_{c, u,v}$ of all three channels $\{Y, Cr, Cb\}$ into the 1D DCT features $\mathbf{T}_i \in \mathbb{R}^{3\times H_P \times W_P}$ of the patch $\mathbf{P}_i$, \emph{i.e.,}
\begin{equation}
    \mathbf{T}_i = \text{FLATTEN}\left(||_{c=Y, u=0, v=0}^{c=Cb,u=H_p,v=W_p}\mathbf{T}^i_{c, u,v}\right),
\end{equation}
where $||$ is the concatenation operation and FLATTEN denotes the feature flatten operation. Therefore, the DCT map $\mathbf{T}$ of the view $\mathbf{I}$ is defined as 
\begin{equation}
    \mathbf{T} = \left[\mathbf{T}_1, \mathbf{T}_2, ..., \mathbf{T}_N \right] \in \mathbb{R}^{(3\times H_P \times W_P) \times (\frac{H}{H_P} \times \frac{W}{W_P})},
\end{equation}
where $N=\frac{H}{H_P} \times \frac{W}{W_P}$ is the number of patches. We reshape $\mathbf{T}$ as the size of $(\frac{H}{H_P}, \frac{W}{W_P})$ with $3\!\times\!H_P\!\times\!W_P$ channels, \emph{i.e.,}
\begin{equation}
    \mathbf{T} \leftarrow \text{Reshape}(\mathbf{T}) \in \mathbb{R}^{(3\times H_P \times W_P) \times \frac{H}{H_P} \times \frac{W}{W_P}}.
\end{equation}

\subsection{Annotation-Denoising Auxiliary Task} ~\label{sec:denoise_task}
Annotation-Denosing auxiliary task recovers the local and fine-grained annotations, \emph{e.g.,} DCT maps and keypoints in our paper, from their noisy latent features with the assistance of image features. Therefore, it aims at enforcing the image backbone to learn fine-grained human part information from different annotations. Concretely, our proposed Annotation-Denosing Auxiliary task can be divided into two steps, \emph{i.e.,} (1) adding noise to latent features of annotations, and (2) recovering the annotations by the decoder with the help of RGB images.

\paragraph{Add Noise to Latent Features of Annotations.} Given one view of image $\mathbf{I}$, its corresponding DCT map $\mathbf{T}$ and keypoints $\mathbf{K}$, we extract their latent features $\mathbf{F}_I$, $\mathbf{F}_T$ and $\mathbf{F}_K$. We  add considerable random noise $\mathcal{U}(-\mu, \mu)$ to the latent features of annotations $\mathbf{F}_T$ and $\mathbf{F}_K$, getting the noisy latent features $\mathbf{F}'_T$ and $\mathbf{F}'_K$. Mathematically, they can be defined as
\begin{equation}
    \mathbf{F}'_T = \mathbf{F}_T + \mathcal{U}(-\mu_T, \mu_T),
    \mathbf{F}'_K = \mathbf{F}_K  + \mathcal{U}(-\mu_K, \mu_K),
\end{equation}
where $\mathcal{U}(-\mu, \mu)$ is the noise with random value at each pixel and $\mu$ as its maximal value. In this paper, we set $\mu_K=\max (\mathbf{F}_K)$  and $\mu_T=\max (\mathbf{F}_T)$ . 

\paragraph{Recover Annotations with the Help of RGB Features.} 
Given the noisy latent features $\mathbf{F}'_T, \mathbf{F}'_K$ and RGB features $\mathbf{F}_I$, we recover the DCT maps $\mathbf{T}$ and keypoints $\mathbf{K}$ from two decoders $\mathcal{D}_T$ and $\mathcal{D}_K$ by using $\mathbf{F}'_T$ and $\mathbf{F}'_K$ as the query features and $\mathbf{F}_I$ as the keys for cross-attention transformer layer, respectively. Mathematically, the recovered features  can be defined as
\begin{equation} \label{eq:recover}
    \mathbf{T}' = \mathcal{D}_T(\mathbf{F}'_T, \mathbf{F}_I), \quad \mathbf{K}' = \mathcal{D}_K(\mathbf{F}'_K, \mathbf{F}_I),
\end{equation}
where $\mathcal{D}_T$ and $\mathcal{D}_K$ is one transformer decoder layer, whose illustration is shown in Fig.~\ref{fig:framework} (b). 

\paragraph{Objective Function.}
Give the actual annotation $\mathbf{T}$ of an image view $\mathbf{I}$ and the recovered features $\mathbf{T}'$ from Eq.~\ref{eq:recover}, the recover loss of DCT maps can be defined by L1 loss as 
\begin{equation} \label{eq:denoi_DCT}
    \mathcal{L}_{de}^{DCT}(\mathbf{T}', \mathbf{T}) = ||\mathbf{T}'-\mathbf{T}||_1.
\end{equation}
Inspired by SimCC~\cite{li2022simcc}, we design the horizontal (denoted by $x$) and vertical (denoted by $y$) coordination classification losses to evaluate the divergence between keypoints and recovered keypoints. Mathematically, the keypoint denoising loss can be formulated as 
\begin{equation} \label{eq:denoi_key}
\begin{aligned}
    \small
    \mathcal{L}_{de}^{kp}(\mathbf{K}', \mathbf{K})\!=\!\sum_{o\in \{x, y\}}D_{KL}\left(\mathcal{H}_o(\mathbf{K}'), \text{OneHot}(\mathbf{K}_o)\right),
\end{aligned}
\end{equation}
where $D_{KL}$ denotes computing Kullback–Leibler divergence, $\mathcal{H}_x: \mathbb{R}^d \rightarrow \mathbb{R}^{kW}$, $\mathcal{H}_y: \mathbb{R}^d \rightarrow \mathbb{R}^{kH}$, and $k>1$ is multiplier used in SimCC. Here, $\text{OneHot}(K_x)$ and $\text{OneHot}(K_y)$ denote the one-hot embedding of the horizontal and vertical coordinates of keypoint annotations $\mathbf{K}$, respectively.

Finally, combining two denoising losses forms the annotation-denoising loss $\mathcal{L}_{de}$, which can be expressed as 
\begin{equation} \label{eq:de}
    \mathcal{L}_{de} = \lambda_1\mathcal{L}_{de}^{kp}(\mathbf{K}', \mathbf{K}) + \lambda_2\mathcal{L}_{de}^{DCT}(\mathbf{K}', \mathbf{K}),
\end{equation}
where $\lambda_1$, $\lambda_2$ are empirical parameters, which are default set to 0.1 and 0.2, respectively.
\subsection{Training and Evaluation} ~\label{sec:training}
\vspace{-1em}
\paragraph{Training.} Apart from our newly designed  denoising loss (Eq.~\ref{eq:de}), our framework is based on MoCo. Therefore, we also integrate the typical contrastive loss as
\begin{equation} \label{eq:contrastive}
    \mathcal{L}_{ctr} = -\log \frac{\exp(\mathbf{F}_I^\top\mathbf{F}'_I/\tau)}{\sum_{i=1}^S\exp(\mathbf{F}_I^\top \mathbf{S}_i/\tau)},
\end{equation}
where $\tau$ is a temperature hyper-parameter, $\mathbf{F}_I=H_q(\mathcal{F}_I(\mathbf{I}))$, $\mathbf{F}_{I'}=H_k(\mathcal{F}_I^m(\mathbf{I}'))$, $\mathcal{F}_I$ is the RGB encoder and $\mathcal{F}^m_I$ is the momentum network. $H_q$ and $H_k$ are 
 multiple-layer  perceptron heads to get global features.
 Here, $\mathbf{S}$ is the  queue and
$\mathbf{S}_i \in \mathbf{S}$ is the $i$-th feature in the queue $\mathbf{S}$ which is iteratively updated by the features of the current mini-batch, and $S$ is the number of features in the queue $\mathbf{S}$.
The parameters $\theta_m$ of $\mathcal{F}^m_I$    are updated by $\mathcal{F}_I$ parameters $\theta_q$ in a momentum manner
as Eq.~\ref{eq:moco}.
\begin{equation}\label{eq:moco}
    \theta_m=\lambda\theta_{m}+(1-\lambda)\theta_{q}, 
\end{equation}
where $\lambda=0.999$  is the momentum coefficient.

Therefore, our total objective function for our Adept is a combination of annotation-denoising loss, \emph{i.e.,} Eq.~\ref{eq:de}, and the contrastive loss, \emph{i.e.,} Eq.~\ref{eq:contrastive}, which can be mathematically formulated as 
\begin{equation} \label{eq:total_loss}
    \mathcal{L} = \mathcal{L}_{ctr} +\mathcal{L}_{de}.
\end{equation}
\paragraph{Evaluation.} For evaluating the effectiveness of our proposed Adept, we only use the image backbone as pretrained model for various downstream tasks, \emph{e.g.,} Person ReID, pose estimation, human parsing, \emph{etc.}, and then finetune all parameters with the data of downstream tasks. We report the performance after finetuning as the metric to evaluate the effectiveness of different pretraining methods.


\subsection{Discussion} \label{sec:discussion}

\paragraph{Relation to Cross-modality Contrastive Methods.} An important stream of human-centric pretraining, such as HCMoCo~\cite{hong2022versatile}, is to contrast between multi-modal data to learn fine-grained information. In this work, we discover that the cross-modality contrastive loss applied in HCMoCo is complementary to our proposed annotation-denoising auxiliary tasks. 
Our experimental results in Tab.~\ref{tab:results} show that incorporating cross-modality contrastive loss ($\mathcal L_{cmc}$\footnote{The exact mathematical expression of $\mathcal L_{cmc}$ is Eq.~1 in~\cite{hong2022versatile}}) among RGB, DCT, and keypoint to our current losses can further reduce EPE by  0.14 on Human3.6M for pose estimation.

\subsection{Method Summary}
The proposed Adept method is summarized in Algorithm \ref{alg:alg1}, which exploits RGB, DCT, and keypoint to assist a human-centric model pretraining.

\begin{algorithm}[!t]
    \caption{Summary of \underline{\textbf{A}}nnotation-\underline{\textbf{de}}noising auxiliary task with key\underline{\textbf{p}}oints and DC\underline{\textbf{T}} maps for human-centric pretraining (Adept)}
    \label{alg:alg1}
    \begin{algorithmic}[1] 
        \REQUIRE Prepared data: two views of a RGB image $\mathbf{I}$ and $\mathbf{I}'$, DCT map $\mathbf{T}$ and keypoints $\mathbf{K}$ of this image.
        Initialized encoder: RGB image encoder $\mathcal{F}_I$, momentum RGB image encoder $\mathcal{F}_I^m$, DCT encoder $\mathcal{F}_T$ and keypoints encoder $\mathcal{F}_K$. 
        Initialized decoder: DCT map and keypoints transformer decoder $\mathcal{D}_T$ and $\mathcal{D}_K$.
        \ENSURE EMA rate $\lambda$, noise $\mathcal{U}$, loss $\mathcal{L}_{ctr}$, $\mathcal{L}_{de}^{DCT}$, $\mathcal{L}_{de}^{kp}$.
        \STATE {\textcolor{gray}{// pre-train the backbone}}
        \STATE {\textbf{$First Stage:$}}
            \FOR {each batch $I$, $I'$ in Dataloader }
                \STATE {$\mathbf{F}_I=\mathcal{F}_I(\mathbf{I})$, $\mathbf{F}_{I'}=\mathcal{F}_{I}^m(\mathbf{I}')$}
                \STATE $\mathcal{F}_I\gets\min\mathcal{L}_{ctr}(\mathbf{F}_I, \mathbf{F}_{I'})$
                \STATE $\mathcal{F}_I=\lambda \mathcal{F}_{I}^m+(1-\lambda)\mathcal{F}_I$
            \ENDFOR
        \STATE {\textbf{$Second Stage:$}}
            \FOR {each batch $I$, $I'$, $T$, $K$ in Dataloader}
                \STATE {$\mathbf{F}_I=\mathcal{F}_I(\mathbf{I})$, $\mathbf{F}_{I'}=\mathcal{F}_{I}^m(\mathbf{I}')$}
                \STATE {$\mathbf{F}_T=\mathcal{F}_T(\mathbf{T})$, $\mathbf{F}_{K}=\mathcal{F}_{K}(\mathbf{K})$}
                \STATE {\textcolor{gray}{// add noise}}
                \STATE {$\mu_K=\max (\mathbf{F}_K)$, $\mu_T=\max (\mathbf{F}_T)$}
                \STATE {$\mathbf{F}'_T = \mathbf{F}_T + \mathcal{U}(-\mu_T, \mu_T)$, $\mathbf{F}'_K = \mathbf{F}_K  + \mathcal{U}(-\mu_K, \mu_K)$}
                \STATE {\textcolor{gray}{// recover annotation from noisy feature}}
                \STATE {$\mathbf{T}' = \mathcal{D}_T(\mathbf{F}'_T, \mathbf{F}_I)$, $\mathbf{K}' = \mathcal{D}_K(\mathbf{F}'_K, \mathbf{F}_I)$}
                \STATE $\mathcal{F}_I, \mathcal{F}_T, \mathcal{F}_K, \mathcal{D}_T$, $\mathcal{D}_K \gets\min\mathcal{L}_{ctr}(\mathbf{F}_I, \mathbf{F}_{I'}) +\min\mathcal{L}_{de}^{DCT}(\mathbf{T}', \mathbf{T}) + \min\mathcal{L}_{de}^{kp}(\mathbf{K}', \mathbf{K})$
                \STATE $\mathbf{F}_I=\lambda \mathbf{F}_{I}^m+(1-\lambda)\mathbf{F}_I$
            \ENDFOR
    \end{algorithmic} 
\end{algorithm}

\section{Experiments}
\subsection{Experiment Setup}
 HRNet-W48~\cite{wang2020deep} are default encoders to capture RGB image and DCT features. COCO~\cite{guler2018densepose} and AIC~\cite{wu2017ai} are the datasets for pretraining. Both of the two datasets provide
paired human RGB images and 2D keypoints, which contain much more diversities than MPII+NTURGBD~\cite{hong2022versatile}. We train the network in two stages. In the first stage, 
the network is trained  with a contrastive loss $\mathcal L_{ctr}$  for 200 epochs and  batch size of 1280. 
In the second stage, 
denoising losses $ \mathcal{L}_{de}$ are added to $\mathcal L_{ctr}$ to train the model for another 110 epochs with  batch size of 256.

\subsection{Implementation Details on Downstream Tasks}
We evaluate our pretrained models on five 
classic human-centric downstream tasks, including pose estimation, human parsing, crowd counting, crowd location, and person ReID on images. 
\begin{itemize}
\item Pose Estimation: For the human pose estimation task, we use the official open-sourced implementation \footnote{https://github.com/facebookresearch/detectron2} and test pose estimation performance on COCO, MPII~\cite{andriluka20142d}, and  Human3.6M~\cite{ionescu2013human3} datasets. We train the network for 130000 iterations with a batch size of 16, a learning rate of 0.01 on 4 NVIDIA V100 GPUs, which takes around 80 hours to train.

\item Human Parsing: We use the official HRNet semantic segmentation implementation \footnote{https://github.com/HRNet/HRNet-Semantic-Segmentation} for the human parsing task. Following implementation instructions, we train the network for 50 epochs with a learning rate of 0.007, with a batch size 40 on 2 NVIDIA V100 GPUs. We use the standard dataset split protocol, where subjects 1, 5, 6, 7, and 8 are for training, and subjects 9 and 11 are for evaluation. Human3.6M offering human part segmentation labels is used to validate our method on the human parsing task.
\item Crowd Counting: ShanghaiTech PartA (SHA) and PartB (SHB) ~\cite{zhang2016single} provide ground truth of crowd number, which aligns with our objective. We use the official HRNet implementation \footnote{https://github.com/taohan10200/IIM} for the crowd human counting task. The batch size is 24 and the max epoch is 200 on two NVIDIA V100 GPUs. The best-performing model on the validation is selected to conduct testing and evaluate our models.
\item Crowd Location: SHA and SHB also supply person location labels, and hence they are adopted as testing datasets to validate crowd location performance. We use the same official HRNet implementation as the previous crowd human counting for this crowd location task and the best-performing model during training is selected to evaluate the model.
\item Person ReID: Market1501~\cite{zheng2015scalable} and MSMT~\cite{wei2018person} datasets are two commonly used datasets to measure ReID method, and we apply them to test our pretrained model on global task with an image-size 256$\times$128. We use the open-sourced HRNet network ReID implementation \footnote{https://github.com/NUISTGY/Person-re-identification-based-on-HRNet} and we train the network for 50 epochs with a learning rate of 0.007, a batch size of 40 on 2 NVIDIA V100 GPUs
\end{itemize}
\subsection{Results of Pretraining on RGB Datasets}
There are a few prior methods that are explored to learn versatile human-centric representation,  and the most related work is HCMoCo. Therefore, we implement HCMoCo on the proposed datasets. Due to the absence of depth which is one modality in HCMoCo, we use DPT-Hybird~\cite{ranftl2021vision} trained on Omnidata~\cite{eftekhar2021omnidata}, to produce pseudo label depth.   Moreover, we propose four state-of-the-art  methods with HRNet-W48 backbone trained on the proposed COCO+AIC datasets for comparisons, which are  MoCo~\cite{chen2020improved}, Dino~\cite{caron2021emerging}, PixPro~\cite{xie2021propagate}, and VICRegL~\cite{bardes2022vicregl}. The backbones of Adept and all other methods are initialized by the pretrained HRNet on ImageNet.

The results of different methods are summarized in Tab.~\ref{tab:results}, where Super IN represents supervised ImageNet (IN) pretrain. It can be observed that the proposed Adept outperforms the state-of-art methods by a large margin on all five tasks under different datasets. 
\begin{table*}[t] \footnotesize
  \centering
  \renewcommand\arraystretch{1.0}
  \caption{Experimental results of our Adept, Adept+$\mathcal L_{cmc}$, and recent state-of-the-art methods on 5 human-centric tasks. Following the most commonly-used metrics, for pose estimation, we report mean Average Precision (mAP) on COCO, PCKh on MPII,  End-Point Error (EPE) on Human3.6M. For 
 Person ReID, we report mAP. For human parsing, we report mean Intersection over Union (mIoU). For crowd counting, we report Mean Absolute Error (MAE). For crowd location, we report F1. We highlight the best result in bold. }
    \begin{tabular}{ccccccccccc}
    \hline
    \multirow{3}[0]{*}{Method} & \multicolumn{3}{c}{Pose Estimation} & Human Parsing  & \multicolumn{2}{c}{Crowd Counting } & \multicolumn{2}{c}{Crowd Location} & \multicolumn{2}{c}{Person ReID} \\ 
          & COCO  & MPII  & H3.6M & Human3.6M & SHA   & SHB   & SHA   & SHB   & Market & MSMT \\
      ~ & (mAP$\uparrow$)&(PCKh$\uparrow$) & (EPE$\downarrow$)  &(mIoU$\uparrow$)& (MAE$\downarrow$ )&(MAE$\downarrow$)&(F1$\uparrow$)&(F1$\uparrow$)&(mAP$\uparrow$)&(mAP$\uparrow$) \\ 
    \hline
    Super IN &  76.7     &  90.8     &   7.36    &   59.32    & 56.85      & 6.05   &   76.3    & 86.2      & 88.1      &53.6  \\
    MoCo~\cite{chen2020improved}  &76.8       &    90.6   &   9.54    & 62.90      &   56.12   &  6.22    & 75.5      & 86.5      &  91.4     &  55.4\\
    Dino~\cite{caron2021emerging}  &   76.7    &   90.8   &    7.15   & 64.36       &   55.97   & 5.64      &    75.5   &  86.6     &  89.7     & 54.8 \\
    PixPro~\cite{xie2021propagate} &   76.7    &    90.9   &   7.16    & 67.25      &  58.47    & 5.82     &   76.4    &  85.2     &  87.2     &48.6  \\
    VicRegL~\cite{bardes2022vicregl} &   76.5    &   90.6   &   7.38    &     62.99   &    55.68    & 5.97     &     76.4  &  85.9     &   87.3    & 48.8 \\
    HCMoCo~\cite{hong2022versatile} &  76.9     &   91.1    &   6.97    &  66.01     &    56.26   & 5.98     &  77.1     &  87.0     &    85.8   & 48.9 \\
Adept (ours) &   77.3    &  92.4    &    6.60   &   71.14    &  53.71     &  5.61&    78.0   &   87.7    &   \textbf{91.5}    & 56.1 \\

 Adept+$\mathcal L_{cmc}$(ours) &   \textbf{77.7}    &  \textbf{93.1}    &    \textbf{6.10}   &   \textbf{71.75}    &  \textbf{52.54}     &      \textbf{5.57} &    \textbf{78.2}   &   \textbf{87.8}    &   \textbf{91.5}    & \textbf{56.2} \\
    \hline
    \end{tabular}%
  \label{tab:results}%
\end{table*}%

\begin{table*}[t] \small
  \centering
    \renewcommand\arraystretch{1.0}
  \caption{Comparison Results between our method and HCMoCo with HRNet-W18 Backbone Trained on MPII+NTURGBD.}
    \begin{tabular}{ccccccccccc}
    \hline
    \multirow{3}[0]{*}{Method} & \multicolumn{3}{c}{Pose Estimation} & Human Parsing & \multicolumn{2}{c}{Crowd Location} & \multicolumn{2}{c}{Person ReID} \\ 
          & COCO  & MPII  & Human3.6M & Human3.6M    & SHA   & SHB   & Market1501 & MSMT \\
      ~ & (mAP$\uparrow$)&(PCKh$\uparrow$) & (EPE$\downarrow$)  &(mIoU$\uparrow$)&(F1$\uparrow$)&(F1$\uparrow$)&(mAP$\uparrow$)&(mAP$\uparrow$) \\ 
    \hline
    Super IN &  73.8    &    90.0  &   9.41    &   56.90         & 75.7      &     84.0  &   \textbf{87.1}    & \textbf{47.90} \\
    CMC~\cite{tian2020contrastive} &  -    &  -    &  -     &   58.93       &    -   &    -   &  -    & - \\
    MMV~\cite{alayrac2020self} &   -   &   -   &   -    &  59.08       &    -   &  -     &   -  & - \\
    HCMoCo~\cite{chen2020improved}  & 73.8      &   90.0    &  8.63     &   62.50         &   75.4    &  84.1     &   78.7    & 28.9 \\
    Adept+$\mathcal L_{cmc}$ (ours) & \textbf{73.8}      &  \textbf{90.3}     &   \textbf{7.82}   &    \textbf{65.08}          &    \textbf{76.4}   &   \textbf{84.5}   &  85.5      & 44.8 \\
    \hline
    \end{tabular}%
  \label{tab:mpii}%
\end{table*}%

\paragraph{Pose Estimation.}
 As shown in Tab.~\ref{tab:results}, our Adept shows superior performance on COCO, MPII, and Human3.6M datasets. Compared with the most competing method, \emph{i.e.,} HCMoCo, our Adept improves the performance on COCO  and MPII datasets by \textbf{+0.4\%}AP and \textbf{+1.3\%}PCKh on MPII respectively. For indoor datasets Human3.6M, our Adept also improves the state-of-the-art method by \textbf{-0.37} EPE. Compared with other state-of-the-art pretraining methods, the superiority of our Adept lies in learning local information from DCT maps. Specifically, compared to MoCo and Dino that only contrast among global images, our Adept designs annotation-denoising auxiliary loss to enforce the image backbone to learn fine-grained semantic information, which could be beneficial to pose estimation tasks. Although HCMoCo design losses to learn fine-grained information from the depth map, our Adept also shows better results for two reasons. First, HCMoCo uses additional pseudo depth maps which may not contain correct semantic information but our Adept learns from a better representation in frequency space. Second, HCMoCo only contrasts features of keypoints and the feature maps of RGB images at the corresponding position. Furthermore, the performance of Adept can be further improved by adding the cross-modality contrastive loss $\mathcal L_{cmc}$ described in Sec.~\ref{sec:discussion} (denoted as Adept+$\mathcal L_{cmc}$), indicating cross-modality contrastive learning can further help the image backbone to extract fine-grained and structural human body information that is beneficial to pose estimation.

\paragraph{Human Parsing.}
It can be observed in Tab.~\ref{tab:results} that our Adept improves the state-of-the-art method, \emph{i.e.,} PixPro, by \textbf{+3.89} mIoU. Interestingly, the performance of HCMoCo that is specifically designed from human-centric downstream tasks is \textbf{-1.24} mIoU worse than fine-grained feature pretraining on natural images, \emph{i.e.,} PixPro, which empirically validates that the pseudo depth map as dense annotations can only provide little or incorrect semantic information which may have a negative influence on dense prediction tasks of human-centric tasks. In contrast, our Adept learns fine-grained semantic information from DCT map, which is a more robust representation that contains more fine-grained semantic information.

\paragraph{Crowd Counting.} Adept and Adept+$\mathcal L_{cmc}$ get the best performance than other state-of-the-art pretraining methods. It reveals the proposed methods learn more information about the global persons for crowd counting than other methods. The importance of global information for the crowd counting task can also be validated by the results that Dino is better than HCMoCo. 

\paragraph{Crowd Location.}
As shown in Tab.~\ref{tab:results}, the proposed Adept performs better than other state-of-the-art pretraining methods both on SHA and SHB datasets. Compared with HCMoCo, Adept obtains \textbf{+0.9} and \textbf{+0.7} gains in terms of F1 score on SHA and SHB, respectively, indicating that our Adept captures  more local information than HCMoCo.

\paragraph{Person ReID.}
We evaluate the performance of our Adept on the person ReID task using Market1501 and MSMT datasets. The proposed Adept surpasses all other pretraining methods. Specifically, Adept obtains significant increases when compared to HCMoCo, VicRegL, and PixPro, which are mainly designed for learning local features.  Compared to global contrastive-based methods, \emph{i.e.,} MoCo and Dino, our Adept+$\mathcal{L}_{cmc}$ further boosts performance by \textbf{+0.1\%} mAP and \textbf{+0.8\%} gains on Market1501 and MSMT, respectively. It can be seen that local information pretraining methods, \emph{e.g.,} PixPro and  HCMoCo, even show worse results than Super IN. This is because person ReID is a global identification task that requires models to learn invariant representation regardless of image transformation. In contrast, MoCo and Dino are able to learn global features, which are invariant to image transformation, resulting in MoCo and Dino achieving better performance. 

\begin{table*}[t]\small
  \centering
  \renewcommand\arraystretch{1.0}
  \caption{Ablation study of annotation-denoising tasks for pose estimation, human parsing, crowd counting, crowd location, and person ReID. }
    \begin{tabular}{cccccccc}
    \hline
    \multicolumn{3}{c}{Loss} & Pose Estimation & Human Parsing & Crowd Counting & Crowd Location & Person ReID \\
    \multirow{2}[0]{*}{$\mathcal{L}_{ctr}$} & \multirow{2}[0]{*}{$\mathcal{L}_{cmc}$} & \multirow{2}[0]{*}{$\mathcal{L}_{de}$} & COCO  & Human3.6M & SHA   & SHA   & Market1501 \\
          &       &       & (mAP$\uparrow$) & (mIoU$\uparrow$) & (MAE$\downarrow$) & (F1$\uparrow$)  & (mAP$\uparrow$) \\
    \hline
    \checkmark & & &76.8& 62.90&56.12&75.5&91.4 \\
    \checkmark & \checkmark& & 76.8      &  65.98     & 55.21     &  76.9   & 91.2\\  
    \checkmark & & \checkmark & \textbf{77.3}      & \textbf{ 71.14}     & \textbf{53.71}     & \textbf{  78.0 }   & \textbf{91.5}\\ 
     \hline
    \end{tabular}%
  \label{tab:ablation1}%
\end{table*}%

\begin{table*}[htbp]\small
  \centering
  \renewcommand\arraystretch{1.0}
  \caption{Ablation study of DCT for pose estimation, human parsing, crowd counting, crowd location, and person ReID.  }
    \begin{tabular}{cccccccc}
    \hline
    \multicolumn{3}{c}{Loss} & Pose Estimation & Human Parsing & Crowd Counting & Crowd Location & Person ReID \\
    \multirow{2}[0]{*}{$\mathcal{L}_{de}$} & \multirow{2}[0]{*}{$\mathcal{L}_{cmc}$} & \multirow{2}[0]{*}{$\mathcal{L}_{DCT}$} & COCO  & Human3.6M & SHA   & SHA   & Market1501 \\
          &       &       & (mAP$\uparrow$) & (mIoU$\uparrow$) & (MAE$\downarrow$) & (F1$\uparrow$)  & (mAP$\uparrow$) \\
    \hline
    \checkmark & \checkmark& &  77.2     &  68.70     &    55.20   &   78.1    & 87.6 \\
    \checkmark & \checkmark& \checkmark & \textbf{77.6}     & \textbf{71.10}      & \textbf{ 52.62}      &    \textbf{ 78.2}  & \textbf{87.7} \\
     \hline
    \end{tabular}%
  \label{tab:ablation2}%
\end{table*}%

\begin{table*}[htbp]\small
  \centering
  \renewcommand\arraystretch{1.0}
  \caption{Ablation study of different frequency domain representations for pose estimation, human parsing, crowd counting, crowd location, and person ReID.}
    \begin{tabular}{cccccc}
    \hline
    Frequency & Pose Estimation & Human Parsing & Crowd Counting & Crowd Location & Person ReID \\
    Domain & COCO  & Human3.6M & SHA   & SHA & Market1501 \\
    Representations & (mAP$\uparrow$) & (mIoU$\uparrow$) & (MAE$\downarrow$) & (F1$\uparrow$)  & (mAP$\uparrow$) \\
    \hline
    DFT & 73.1 & 64.60 & 57.30 & 75.0 & 84.8 \\
    DCT & \textbf{77.6} & \textbf{71.10} & \textbf{ 52.62} & \textbf{ 78.2} & \textbf{87.7} \\
     \hline
    \end{tabular}%
  \label{tab:ablation_dct_dft}%
\end{table*}%

\paragraph{Summary.}
Through the observations in Tab.~\ref{tab:results}, we find that MoCo and Dino benefit global tasks, \emph{e.g.,} Person ReID, but show little improvement on local tasks, \emph{e.g.,} pose estimation and crowd location. Local pretraining methods, \emph{e.g.,} PixPro and HCMoCo, force the backbone to capture local features, which is helpful for local tasks.
Our Adept incorporates local and global learning into the unified framework by the typical contrastive loss and the newly designed annotation-denoising auxiliary tasks, which could help the backbone to learn local and global information simultaneously. In addition, we find adding the cross-modality loss $\mathcal L_{cmc}$ can further improve the performance of our Adept on pose estimation, human parsing, crowd location, and crowd counting tasks while maintaining the performance on person ReID.

\subsection{Results of Pretraining on MPII+NTURGBD Datasets with Depth Data}
To make a direct comparison with HCMoCo, we also pretrain our backbone on MPII+NTURGBD datasets by our proposed method, and test our method on pose estimation, human parsing, crowd location, and person ReID. For fair comparisons with HCMoCo, the encoders for image and DCT used in this section are HRNet-W18.

Tab.~\ref{tab:mpii} lists the results of different methods using MPII+NTURGBD as pretraining datasets. Adept+$\mathcal L_{cmc}$ surpasses or is comparable to HCMoCo on all tasks with different datasets. To be concrete, Adept+$\mathcal L_{cmc}$ achieves \textbf{+0.3\%} PCKh gains on MPII, \textbf{-0.81} EPE gains on Human3.6M for pose estimation, \textbf{+2.58} mIoU increase on Human3.6M for human parsing, and \textbf{+1.0} F1 score and \textbf{+0.4} F1 score on SHA and SHB for crowd location. These results further prove our Adept+$\mathcal L_{cmc}$ learns more useful local features than HCMoCo by annotation-denoising task. For person ReID, Adept+$\mathcal L_{cmc}$ shows a significant increase than HCMoCo by \textbf{+6.8\%} and \textbf{+15.9\%} mAP on Market1501 and MSMT, respectively, which indicates our method learns more robust global representation.  

We can see that Super IN shows higher mAP than Adept+$\mathcal L_{cmc}$ for ReID tasks. The reason is that  NTURGBD provides very limited identities (only 11 identities), where almost every identity contains more than one thousand images with varying gestures. When performing contrastive learning on RGB images, it considers other images in the current mini-batch as negative samples, while most of such images may belong to the same identity, causing a significant performance drop on ReID tasks. Therefore, NTURGBD dataset is not suitable for pretraining ReID tasks, and the large-scale COCO and AIC datasets, which contain much more identities, exhibit the potential to be better pretraining datasets for human-centric downstream tasks, including both global and local information.

\subsection{Ablation Study}
To verify the effectiveness of key components of our method,  we perform a thorough ablation study with COCO and AIC as pretraining datasets.

\noindent \textbf{Effectiveness of annotation-denoising auxiliary task.}
We observe in Tab.~\ref{tab:ablation1}   that Adept which merges denoising loss $\mathcal{L}_{de}$ and contrastive losses  $\mathcal{L}_{ctr}$ outperforms  
MoCo only using $\mathcal{L}_{ctr}$  by \textbf{+0.5\%} mAP for pose estimation, \textbf{+8.24} mIoU for  human parsing, \textbf{-1.41} MAE  for crowd counting, \textbf{+2.2} F1 score for crowd location, and keep comparable performance for ReID task. These results demonstrate that the proposed annotation-denoising auxiliary task is able to learn  local and fine-grained features, favorable to dense tasks. 
Compared with the cross-modality contrastive loss $\mathcal{L}_{cmc}$ used in HCMoCo, the annotation-denoising loss $\mathcal{L}_{de}$ used in Adept is further better on all 5 tasks. Specifically, Adept brings \textbf{+0.5\%} mAP improvement for pose estimation, \textbf{+5.16} mIoU increase for  human parsing, \textbf{+0.8} F1 increase  for crowd location, \textbf{+0.3} mAP  improvement for person ReID, and reduce \textbf{-0.5} MAE  or crowd counting, which signals that our annotation-denoising auxiliary task is more powerful than HCMoCo to capture rich local features while reserving more invariant information for the global task.

  


\noindent \textbf{Effectiveness of DCT.}
As shown in Tab.~\ref{tab:ablation2}, 
by adding losses related to DCT $\mathcal{L}_{ctr}$ , the performance shows significant improvements  for pose estimation, human parsing, and crowd counting, which validates that frequency information is essential to human-centric pretraining. This is because DCT provides complementary information to images, such as semantic patterns, for human-centric perception tasks. Therefore, studying the fusion method of DCT and image in downstream tasks is worthy of exploration in future work. 

\begin{figure}[t]
    \centering
    \includegraphics[width=0.8\linewidth]{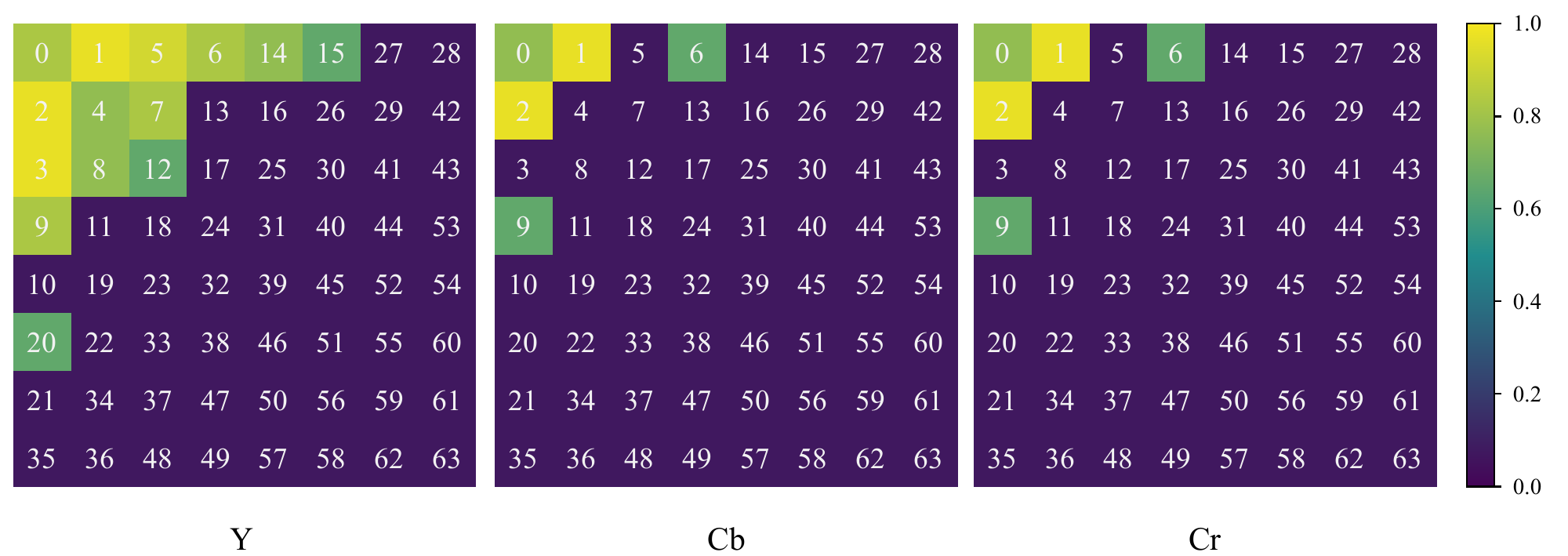}
    \caption{Visualization of heat maps for Y, Cb, and Cr components on the COCO validation dataset.}
    \label{fig:heatmap}
\end{figure}

\noindent \textbf{Further exploration of different frequency domain representations}. To further explore the advantages of DCT maps in this annotation-denoising auxiliary task, we replace DCT maps with Discrete Fourier Transform (DFT) maps and conduct experiments to assess their effectiveness. The experimental results (see Tab.~\ref{tab:ablation_dct_dft}) demonstrate that DCT maps significantly enhance the performance on human-centric perception tasks with +2.9\% AP to +6.5\% mIoU across five downstream tasks compared to DFT maps, validating the superiority of DCT maps in training network models. For an in-depth analysis, we followed the method from~\cite{xu2020learning} and visualize the probability heatmap of different frequency channel selections on the COCO validation set. As shown in Fig.~\ref{fig:heatmap}, low-frequency channels (boxes with small indices) are selected much more frequently than high-frequency channels (boxes with large indices), indicating that low-frequency channels are generally more informative for vision perception tasks and better support the network in learning semantic information. Compared to DFT, DCT basis consists solely of cosine waves, which concentrate low-frequency components, making it more effective in separating high-frequency and low-frequency components. Therefore, DCT maps, due to their enhanced low-frequency channel concentration, are more effective for network learning of representational information.

\begin{table*}[htbp]\small
  \centering
  \renewcommand\arraystretch{1.0}
  \caption{Performance of Adept on pose estimation, human parsing, crowd counting, crowd location, and person ReID tasks when scaling up noise level.  Style: \textbf{best}, \underline{second}.}
    \begin{tabular}{cccccc}
    \hline
    \multirow{3}[0]{*}{Noise Level} & Pose Estimation & Human Parsing & Crowd Counting & Crowd Location & Person ReID \\
     & COCO  & Human3.6M & SHA   & SHA   & Market1501 \\
     & (mAP$\uparrow$) & (mIoU$\uparrow$) & (MAE$\downarrow$) & (F1$\uparrow$)  & (mAP$\uparrow$) \\
    \hline
    0.0× & 76.8 & 62.90 & 56.12 & 75.5 & 91.4 \\
    0.1× & 76.9 & 64.60 & 57.30 & 75.0 & 91.4 \\
    0.3× & 77.1 & \underline{71.10} & 55.67 & 76.0 & \underline{91.5} \\
    1× & \underline{77.3} & \textbf{71.14} & \textbf{53.71} & \underline{78.0 } & \underline{91.5} \\
    3× & \textbf{77.6} & 68.8 & \underline{54.22} & \textbf{78.2} & \textbf{91.6} \\
    10× & 77.0 & 67.53 & 55.01 & \textbf{78.2} & 91.3 \\
     \hline
    \end{tabular}%
  \label{tab:ablation_noise_level}%
\end{table*}%
\noindent \textbf{Impact of Noise Injection}
To assess the impact of different noise levels in the annotation-denoising auxiliary task, we trained Adept on ablation noise levels comprising 0.0×, 0.1×, 0.3×, 1×, 3×, and 10× noise levels. The results in Tab.~\ref{tab:ablation_noise_level} indicate that performance across all tasks initially increases with higher noise levels, reaching a peak before declining as the noise becomes too excessive. While the optimal noise level varies slightly across tasks, a noise level of 1× generally yields the best or second-best performance across all five tasks. For instance, at this level, Human Parsing achieves the highest performance (\textbf{71.14\% mIoU}), and Crowd Counting also reaches its peak (\textbf{53.71\% MAE}). Meanwhile, Pose Estimation (\underline{77.3\% mAP}), Crowd Location (\underline{78.0\% F1}), and Person ReID (\underline{91.5\% mAP}) record their second-best performances. The experimental results also reveal that varying noise levels have a relatively small impact on Person ReID, whereas Human Parsing, a local task, is more significantly affected.

\begin{table*}[htbp]\small
  \centering
  \renewcommand\arraystretch{1.0}
  \caption{Comparison Results between Visual Prompt Tuning (VPT) and Full Finetuning (FT) for pose estimation, human parsing, crowd counting, crowd location, and person ReID.  Style: \textbf{best}.}
    \begin{tabular}{ccccccc}
    \hline
    \multirow{3}[0]{*}{Method} & & Pose Estimation & Human Parsing & Crowd Counting & Crowd Location & Person ReID \\
     & & COCO  & Human3.6M & SHA   & SHA   & Market1501 \\
     & & (mAP$\uparrow$) & (mIoU$\uparrow$) & (MAE$\downarrow$) & (F1$\uparrow$)  & (mAP$\uparrow$) \\
    \hline
    \multirow{3}[0]{*}{VPT} & Super IN & 56.8 & 45.30 & 65.17 & 76.6 & 71.4 \\
    & HCMoCo~\cite{chen2020improved} & 65.9 & 64.62 & 54.30 & 77.4 & 81.3 \\
    & Adept (ours) & 71.1 & 58.93 & \textbf{52.77} & \textbf{79.1} & 82.6 \\
    \hline
    \multirow{3}[0]{*}{FT}& Super IN & 74.9 & 61.07 & 65.12 & 76.0  & 88.2 \\
    & HCMoCo~\cite{chen2020improved} & 75.7 & 71.0 & 54.22 & 77.2 & 90.3 \\
    & Adept (ours) & \textbf{77.3} & \textbf{71.14} & 53.71 & 78.0 & \textbf{91.5} \\
     \hline
    \end{tabular}%
  \label{tab:ablation_vpt_ft}%
\end{table*}%

\noindent\textbf{Full Finetuning versus Prompt Tuning.} To investigate the performance of the pre-trained models using different parameter learning strategies on downstream tasks, we reproduce the VPT method~\cite{han2024facing} and conduct ablation experiments for both VPT and FT methods. As shown in Tab.~\ref{tab:ablation_vpt_ft}, our Adept method demonstrates performance advantages under both optimization strategies. Specifically, under VPT training, our Adept method achieves the best performance on Crowd Counting and Crowd Location tasks, with an MAE of \textbf{52.77}\% and an F1 score of \textbf{79.1}\%. We analyze that there exists a substantial gap between the pretraining process and the Crowd Counting and Crowd Location downstream tasks, making FT more prone to overfitting due to its numerous tunable parameters. In contrast, for Pose Estimation, Human Parsing, and Person ReID tasks, the pretraining process includes similar learning objectives, resulting in samll task disparity, making FT less susceptible to overfitting. Consequently, our Adept method outperformed VPT with FT, exceeding it by \textbf{+6.2}\% mAP, \textbf{+12.21}\% mIoU, and \textbf{+8.9}\% mAP, respectively.

\begin{figure*}
    \centering
    \includegraphics[width=\linewidth]{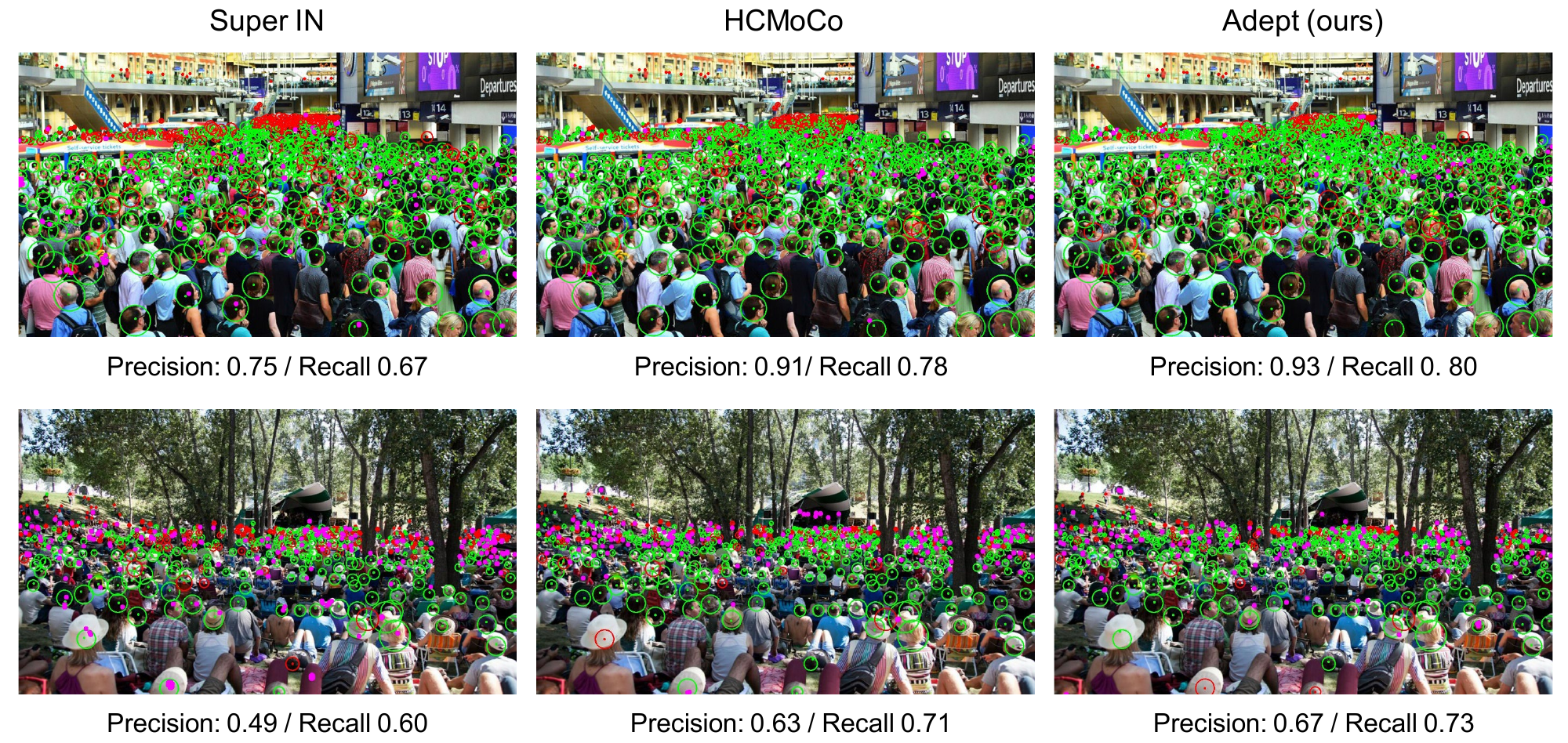}
    \caption{Some typical visualization results of our method Adept, baseline line method (ImageNet Pretrain), and the compared HCMoCo on SHHA validation set. The green, red and magenta points denote true positive (TP), false negative (FN), and false positive (FP), respectively. For easier reading, images are labeled with the precision and recall rates.}
    \label{fig:vis-location}
\end{figure*}
\begin{figure*}
    \centering
    \includegraphics[width=\linewidth]{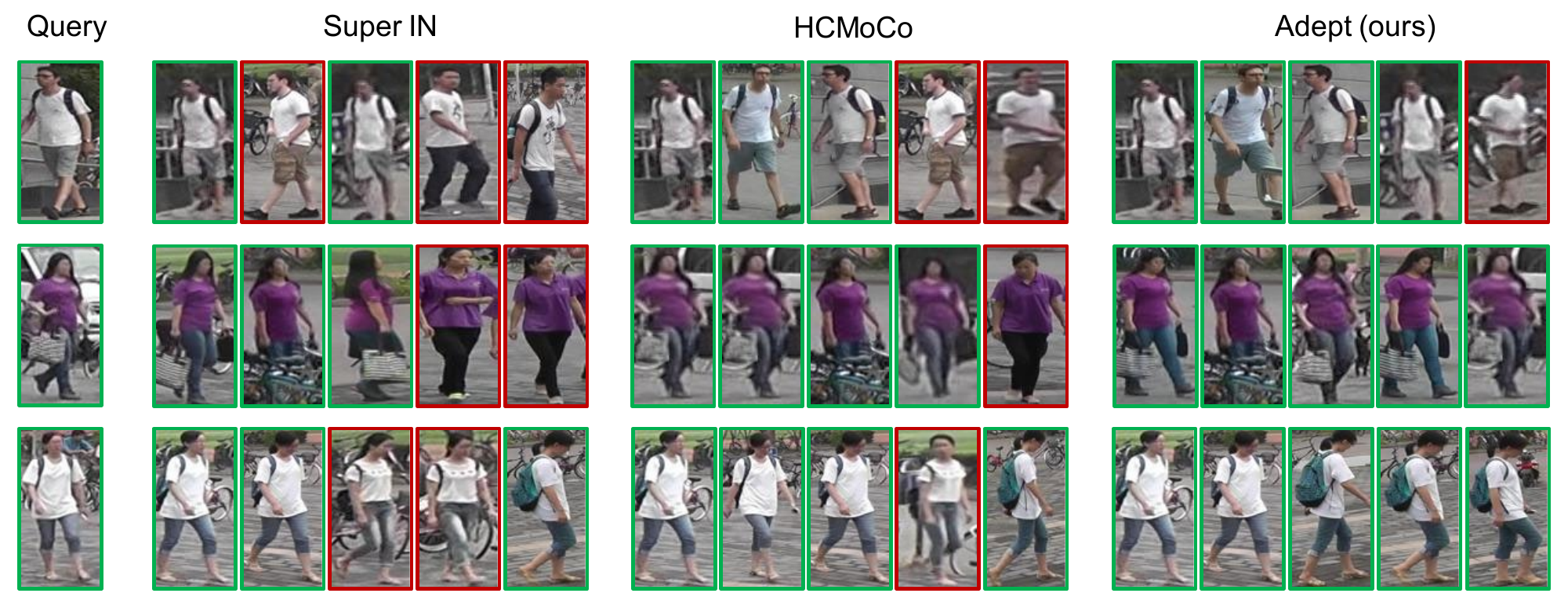}
    \caption{Illustration of some typical visualization results of our method Adept, baseline line method (ImageNet Pretrain), and the compared HCMoCo on Market1501 dataset. For easier reading, green and red boxes mean true and false matches.}
    \label{fig:vis-reid}
\end{figure*}

\subsection{Qualitative Results}
To verify the effectiveness of our human-centric pretraining method, we provide a qualitative comparison of RGB visualization results on \textit{Crowd Counting}, \textit{Crowd Localization}, \textit{Person Re-Identification}, and \textit{Pose Estimation} tasks as examples.

To intuitively demonstrate the localization and counting results, we compare two baseline methods (ImageNet pretrain model and HCMoCo pretrain model) and the proposed Adept as shown in Fig.~\ref{fig:vis-location}, which illustrates two groups of typical samples from the SHHA validation set. For clearer interpretation, true positive results are marked with green points, false negative results are indicated with red points, and false positive results are marked with magenta points. For both examples, Adept demonstrates better accuracy and recall. False negatives mainly occur in areas where the visibility of individuals is poor due to high density and low pixel quality, as seen in the upper region of the first example. False positives are primarily found in regions with blurred backgrounds and foregrounds, such as the middle area of the second example.

We visualize the retrieval results of our method Adept, baseline line method ImageNet Pretrain, and the compared HCMoCo in Fig.~\ref{fig:vis-reid}. Given a query image, Adept demonstrates the best accurate retrieval results. We select the top 5 retrieval results and use green boxes to indicate true matches and red boxes to indicate false matches. For the third example, the 3\textsuperscript{rd} image in the ImageNet pretrain method and the 4\textsuperscript{th} image in the HCMoCo method show false matches with similar clothing and posture to the query, but Adept avoids this interference. This indicates that our pretraining method better enables the model to accurately perceive detailed features in the images.

\begin{figure*}
    \centering
    \includegraphics[width=\linewidth]{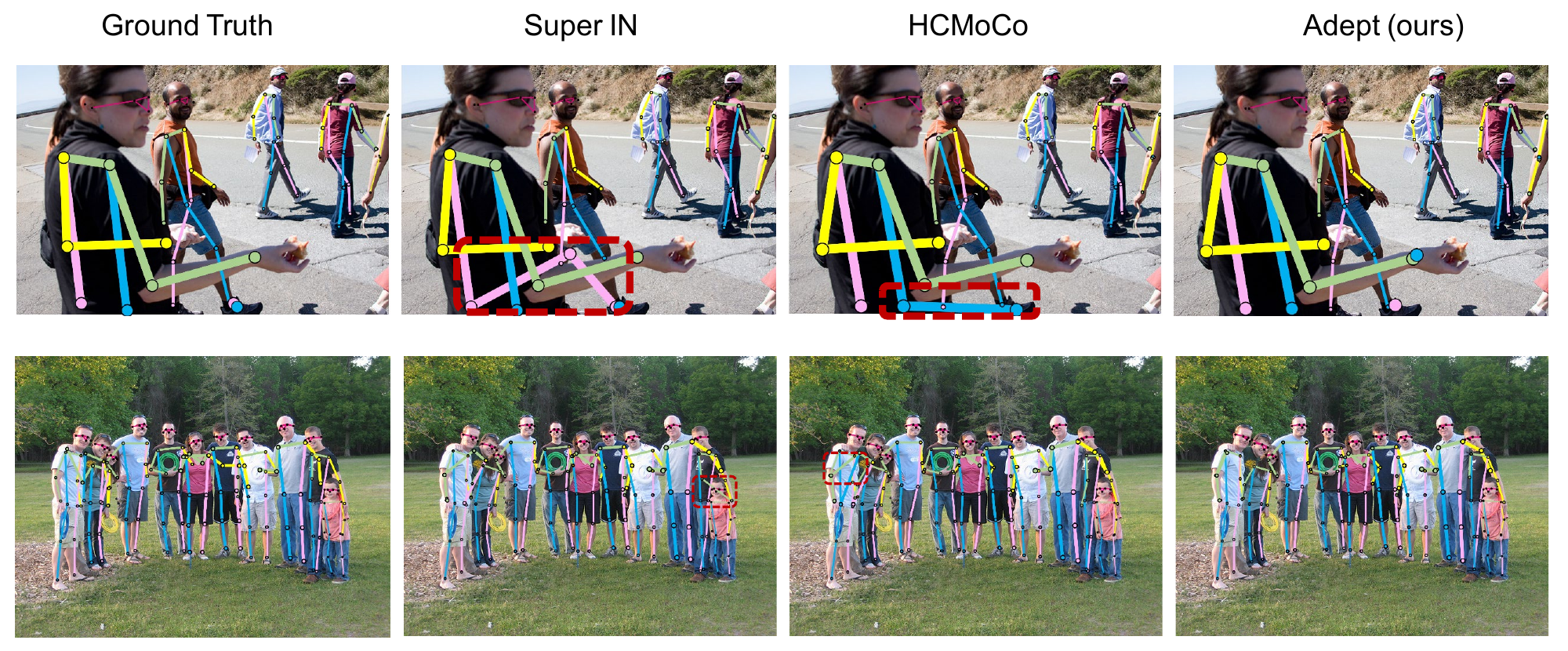}
    \caption{Some typical visualization results of our method Adept, baseline line method (ImageNet Pretrain), and the compared HCMoCo on COCO human pose estimation test set. We select representative images with various human sizes, multiple persons, different poses. To facilitate comparison with the ground truth labels, we have highlighted the incorrect recognition results with red dashed boxes}
    \label{fig:vis-poseSuper}
\end{figure*}

We also visualize the pose estimation results of our method on the MS COCO dataset. As shown in Fig.~\ref{fig:vis-poseSuper}, Adept can generate accurate pose estimation results in challenging cases involving heavy occlusion, various postures, and different scales, demonstrating its strong representation ability. Incorrect recognition results are highlighted with red dashed boxes in the two examples. Adept avoids interference from other individuals, as seen in the first example with the ImageNet pretrain method, and errors due to occlusion, as observed in the first example with HCMoCo. The results from Adept are highly consistent with the ground truth.


\section{Conclusion and Limitations}
\noindent\textbf{Conclusion.} In this work, we propose the  versatile multi-modal pre-training framework Adept, specifically designed for local and global human-centric perception tasks at the same time. Annotation-denoising auxiliary task is designed with a DCT map and keypoint to learn fine-grained local features. We show that introducing a denoising loss to global loss significantly improves the performance on local tasks, for example, human parsing and pose estimation while preserving the recognition accuracy. Extensive experiments on five different human-centric downstream tasks verify the effectiveness of our pretraining framework for global and local image understanding.

\noindent\textbf{Limitations.} Currently, due to limitations in training data, \emph{e.g.,} limited identity information, Adept exhibits a relative bias in performance improvements across the five downstream tasks. Specifically, Adept shows more pronounced performance improvements on local tasks, \emph{e.g.,} human parsing and pose estimation, while demonstrating relatively lower performance gains on ReID tasks. Additionally, the scope and capabilities of Adept are constrained by resource limitations and training costs, preventing further exploration into more generalizable human-centric tasks, such as attribute recognition, skeleton, and 3D pose estimation. Overcoming these barriers will require substantial future work and will help us fully realize the theoretical potential of Adept.

\appendix



\bibliographystyle{model1-num-names}

\bibliography{cas-refs}



\vspace{10em}
\bio{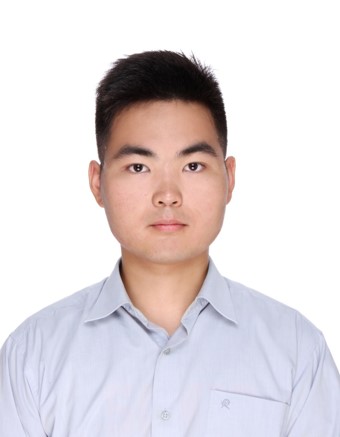}{Weizhen He} received the B.S.degree in College of Electrical Engineering from Zhejiang University, Hangzhou, China, in 2021. He is currently working toward the Ph.D. degree in the major of Control Theory and Control Engineering in Zhejiang University. His research interests include human-centric artificial intelligence, person re-identification and object detection.
\vspace{0em}
\endbio

\bio{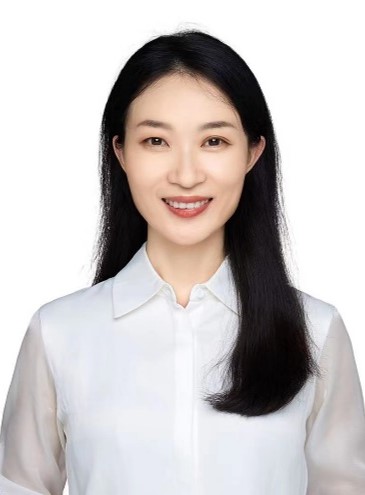}{Yunfeng Yan} received the Ph.D. degree in Electrical Engineering from Zhejiang University, Hangzhou, China, in 2019. She is currently an Associate Research Fellow with the College of Electrical Engineering, Zhejiang University, Hangzhou, China. Her research interests include computer vision, operation situational awareness, and abnormality monitoring of distributed generation equipment.
\vspace{0em}
\endbio

\bio{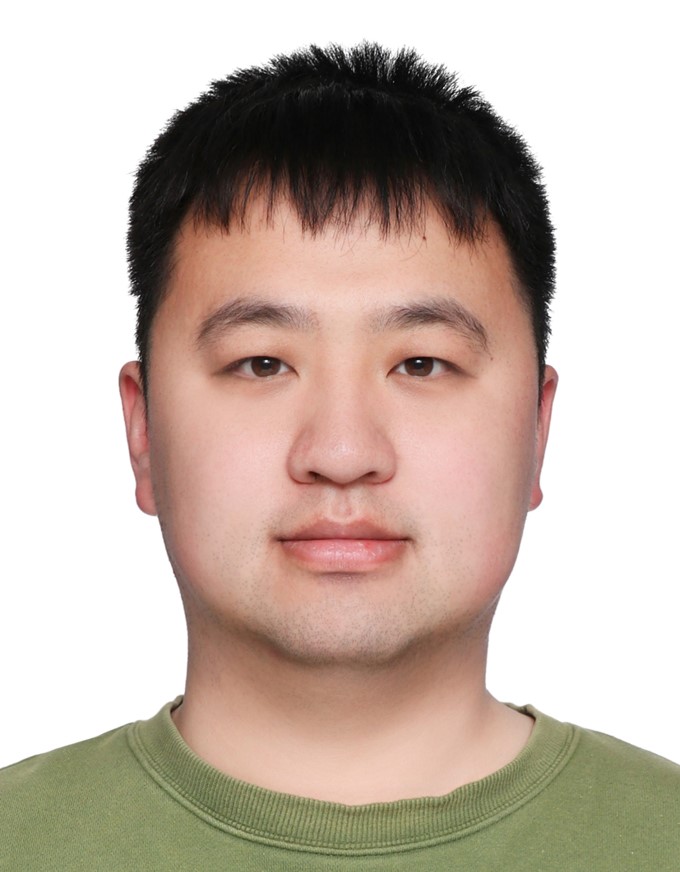}{Shixiang Tang} received the Ph.D degree from the University of Sydney. Prior to that, he received the Master of Philosophy from the Chinese University of Hong Kong in 2018 and Bachelor of Science from Fudan University. His interests lie in machine learning and computer vision, especially self-supervised learning and foundation models. He has published about 10 papers in top-tier conferences and journals, e.g., CVPR, ICCV, NeurIPS, Nature Physics and Nature Materials.
\vspace{1em}
\endbio

\bio{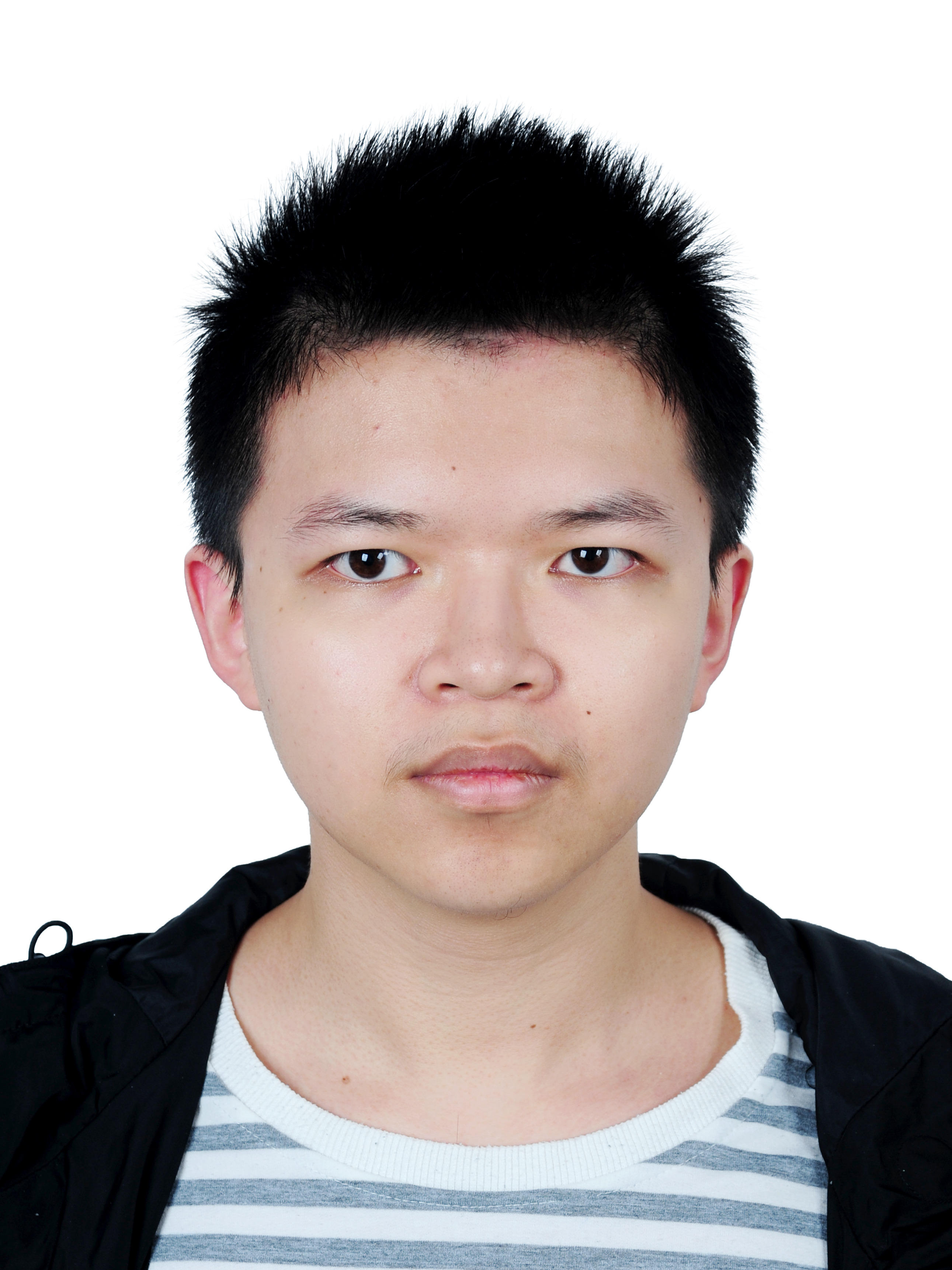}{Yiheng Deng} received the B.S.degree in School of Electrical Engineering from Chongqing University, Chongqing, China, in 2018. He is currently working toward a Ph.D. degree in the major of electrical engineering at Zhejiang University, Hangzhou, China. His research interests include deep learning, person re-identification and object detection.
\vspace{2em}
\endbio

\bio{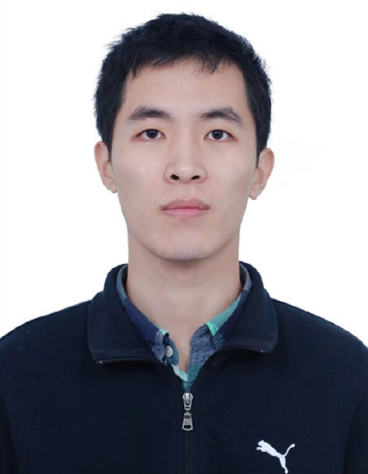}{Yangyang Zhong} received the Master's degree in Control Engineering from Northeast University, Shenyang, China. He is currently pursuing his Ph.D. degree in Electronic Information at Zhejiang University, Hangzhou, China. He is also a member of the Chinese Association for Artificial Intelligence (CAAI). His research interests include large language models, emotional intelligence, computer vision, and natural language processing.
\endbio

\bio{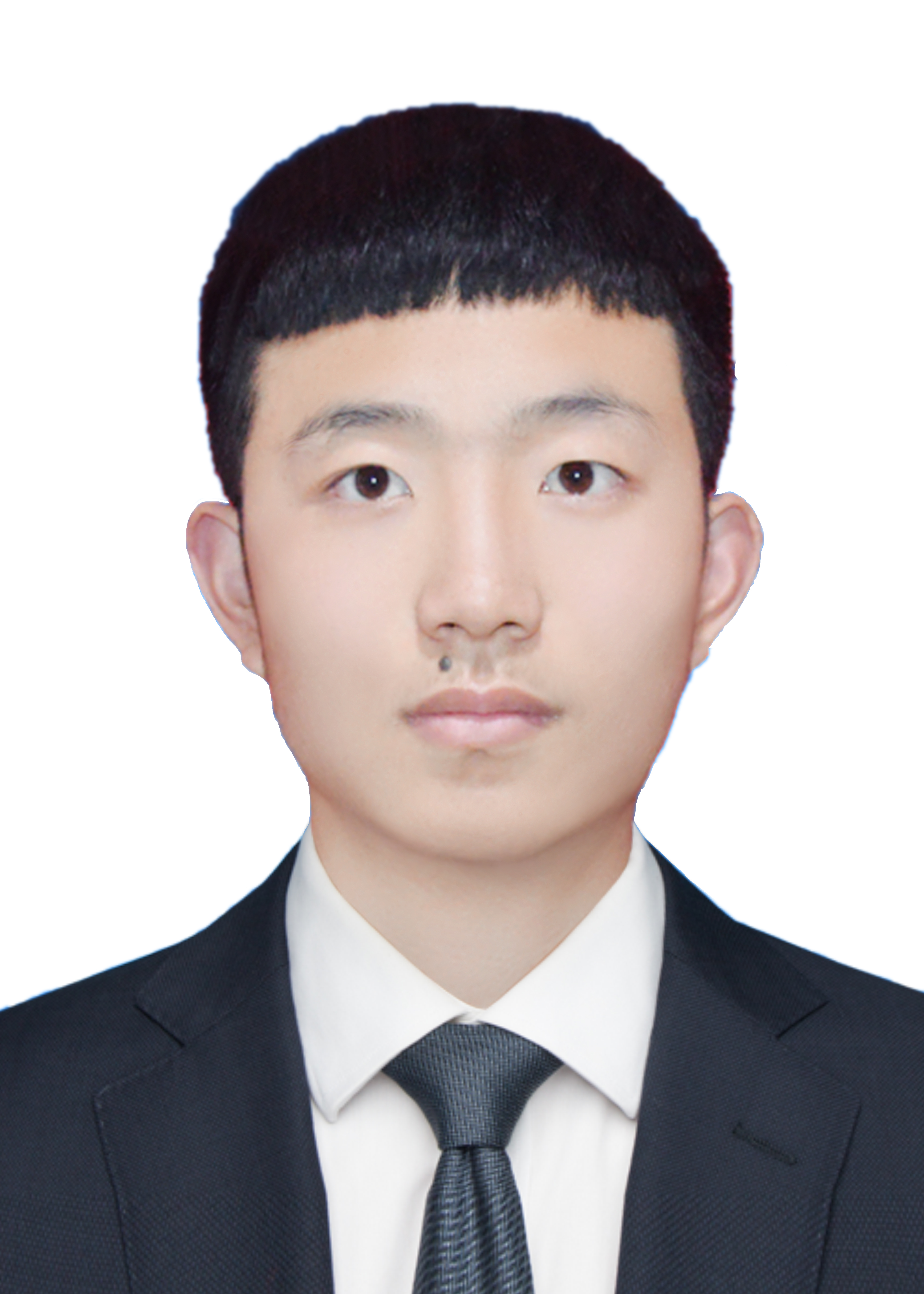}{Pengxin Luo} received the degree of bachelor from the Northwest A$\&$F University in 2023. Currently, he is currently working towards the M.E. degree in Electronic Information from Zhejiang University. His current research direction includes deep learning and computer vision.
\vspace{4em}
\endbio

\bio{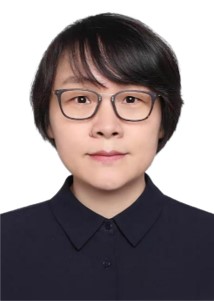}{Donglian Qi} received the Ph.D. degree from the School of Electrical Engineering, Zhejiang University, China, in 2002. She is currently a Full Professor and a Ph.D. Advisor with Zhejiang University. Her recent research interest covers intelligent information processing, chaos systems, and nonlinear theory and application.
\endbio

\end{document}